\documentclass{article}

\usepackage{PRIMEarxiv}

\usepackage[utf8]{inputenc} 
\usepackage[T1]{fontenc}    
\usepackage{hyperref}       
\usepackage{url}            
\usepackage{booktabs}       
\usepackage{amsfonts}
\usepackage{float}
\usepackage{nicefrac}       
\usepackage{microtype}      
\usepackage{lipsum}
\usepackage{fancyhdr}       
\usepackage{graphicx}       
\graphicspath{{media/}}     
\usepackage{multirow}
\usepackage{multicol}
\usepackage{bm}
\usepackage{overpic}
\usepackage{cite}
\usepackage{subcaption}
\usepackage{amsmath}
\usepackage{amssymb}
\usepackage{comment}
\usepackage{algorithm}
\usepackage{algorithmic}
\usepackage{color}
\usepackage{makecell}
\newcommand{\vect}[1]{\boldsymbol{#1}}

\title{Robustness Evaluation of Offline Reinforcement Learning for Robot Control Against Action Perturbations}

\author{
  Shingo Ayabe, Takuto Otomo \\
   Graduate School of Science and Engineering \\
   Chiba University \\
   Chiba, Japan \\
  \texttt{\{ayabe.shingo, takutootomo\}@chiba-u.jp} \\
   \And
  Hiroshi Kera\\
  Graduate School of Informatics \\
  Chiba University \\
  Chiba, Japan \\
  Zuse Institute Berlin \\
  Berlin, Germany \\
  \texttt{kera@chiba-u.jp} \\
  \And
  Kazuhiko Kawamoto \\
  Graduate School of Informatics \\
  Chiba University \\
  Chiba, Japan \\
  \texttt{kawa@faculty.chiba-u.jp} \\
}

\begin{document}
\maketitle

\begin{abstract}
Offline reinforcement learning, which learns solely from datasets without environmental interaction, has gained attention. This approach, similar to traditional online deep reinforcement learning, is particularly promising for robot control applications. Nevertheless, its robustness against real-world challenges, such as joint actuator faults in robots, remains a critical concern. This study evaluates the robustness of existing offline reinforcement learning methods using legged robots from OpenAI Gym based on average episodic rewards. For robustness evaluation, we simulate failures by incorporating both random and adversarial perturbations, representing worst-case scenarios, into the joint torque signals. Our experiments show that existing offline reinforcement learning methods exhibit significant vulnerabilities to these action perturbations and are more vulnerable than online reinforcement learning methods, highlighting the need for more robust approaches in this field.
\end{abstract}


\section{Introduction}
\label{sec:intro}
Offline reinforcement learning (offline RL)~\cite{offline_rl_1} has emerged as a novel approach across various domains like healthcare~\cite{offline_rl_hc}, energy management~\cite{offline_rl_em}, and robot control~\cite{offline_rl_rc}. 
Unlike online RL, which requires direct interaction with environments, offline RL learns solely from pre-collected datasets without environmental interaction.
Hence, offline RL can potentially reduce costs and risks associated with frequent environmental interactions.  

For offline RL to be more widely applicable, robustness is an essential requirement. 
The robustness of offline RL can be broadly categorized into two types: training-time robustness~\cite{robust_trngtime_1,robust_trngtime_2,robust_trngtime_3} and testing-time robustness~\cite{robust_testtime_1,robust_testtime_2,robust_testtime_3,robust_testtime_4,RORL,MICRO,robust_testtime_5}.
Training-time robustness focuses on effectively training policies with corrupted datasets, which may include perturbations caused by data logging errors and intentional data augmentation~\cite{robust_trngtime_1,corruption_1}.
Testing-time robustness, on the other hand, refers to the ability to handle discrepancies between training and deployment environments. In robot control, these discrepancies include state-space perturbations (e.g., sensor perturbations)~\cite{state_perturbation_1,state_perturbation_2} and action perturbations (e.g., mismatched friction coefficients or actuator failures)~\cite{adv_trng_1,o2022_adv,action_perturbation_1}.
Online RL addresses similar challenges, often referred to as the reality gap, through methods like
domain randomization~\cite{domain_random_1,domain_random_2,domain_random_3} and adversarial training~\cite{adv_trng_1,adv_trng_2,adv_trng_3,adv_trng_4,adv_trng_5}.
However, whether these solutions can be effectively applied to offline RL remains unclear.

Online RL benefits from extensive exploration across diverse state-action pairs via direct environmental interactions, enabling broad coverage of the state-action space.
In contrast, offline RL is restricted to exploring only
the state-action pairs available in the pre-collected training dataset, rendering its performance highly dependent on the dataset's quality and coverage.
Moreover, offline RL relies on conservative training to mitigate the overestimation of action values caused by out-of-distribution actions~\cite{offline_rl_2}.

Existing studies~\cite{RORL,MICRO,robust_testtime_5} on testing-time robustness in offline RL have primarily focused on state-space perturbations.
Robustness against state-space perturbations is typically achieved by minimizing the sensitivity of the policy to small input variations. This relies on the assumption that state-space perturbations should not lead to drastically different actions. For example, robustness can be improved by incorporating a smoothing loss~\cite{RORL}, which encourages the policy's output distribution to change smoothly in response to small state-space perturbations, or regularization techniques such as KL divergence constraints that align outputs with and without perturbations~\cite{robust_testtime_5}.
However, in real-world scenarios, especially in robotics, action perturbations, such as actuator failures, are equally critical but remain unexplored.
Action-space perturbations are directly added to the actions produced by the policy network, and thus affect the environment's dynamics and the resulting rewards. As a result, 
these perturbations lead to deviations from the intended behavior dictated by the policy network, posing significant challenges to the reliability of offline RL systems.
Therefore, the policy must output actions inherently robust to such perturbations, still achieving high rewards despite them.
The distinction between robustness to state-space perturbations and action-space perturbations is particularly critical in offline RL. Offline RL algorithms typically impose conservative constraints to align the learned policy closely with the dataset's behavior policy~\cite{offline_rl_1}. 
Therefore, successful learning under action perturbations requires the dataset to include actions that are already adapted to such perturbations, along with corresponding rewards and next states. Collecting such data is typically impractical, underscoring the challenge of studying robustness to action-space perturbations in offline settings.

This study investigates the testing-time robustness of existing offline RL methods~\cite{BCQ,TD3+BC,IQL} against random and adversarial action perturbations.
Various adversarial attack strategies have been proposed in the context of online RL~\cite{adv_trng_1,adv_trng_4,adv_trng_5}, such as gradient-based and curriculum-based methods. 
However, these methods typically require the adversarial perturbations to be jointly optimized with the policy, assuming interactive access to the environment. 
This joint optimization is challenging to apply in offline settings, where the policy must be learned solely from fixed datasets.
To address this limitation, differential-evolution-based perturbations~\cite{o2022_adv} are adopted in this study as a representative method that does not rely on joint policy optimization, making it suitable for offline RL.
Specifically, to evaluate the robustness of these methods, we simulate actuator failures by perturbing the actions generated by the policy network.
These offline RL methods limit exploration to remain within the training data distribution, and we hypothesize this limitation renders them vulnerable to unencountered perturbations.
This vulnerability is particularly critical in the case of adversarial perturbations, which are intentionally crafted to exploit the system under worst-case failure scenarios.
Adversarial perturbations underscore the need to understand their effects to develop reliable control systems. 
However, the impact of action-space perturbations on the robustness in offline RL remains yet to be fully explored.

Our approach involves simulating operational challenges within the MuJoCo physical simulation environment~\cite{MuJoCo}, focusing on forward walking tasks. 
We evaluate the testing-time robustness of offline RL methods using the average episodic reward.
Experimental results reveal that the existing methods are highly vulnerable to action perturbations.
Furthermore, inspired by typical defensive strategies in online RL---training in environments with added perturbations—--we train policies on datasets augmented with action perturbations and evaluate their robustness. The results, however, demonstrate no significant improvement, highlighting the need for more robust offline RL methods to address failures effectively.
The contributions of this paper are as follows:
\begin{itemize}
    \item We evaluate the testing-time robustness of the offline RL methods against random and adversarial action perturbations and reveal that the offline RL models are more vulnerable to action perturbations than online RL.
    \item We find that the testing-time robustness of the offline RL models depends on the state-action coverage of the training dataset.
    \item We show that training with action-perturbed datasets alone does not improve the testing-time robustness of offline RL against action perturbations.
\end{itemize}

\section{Related work}
\label{sec:2}
\subsection{Offline reinforcement learning}
\label{sec:2-1}
In offline training, Q-values tend to be overestimated for unseen state-action pairs that are not represented in the dataset~\cite{offline_rl_1}. Since this overestimation cannot be corrected through direct interaction with the environment, traditional online RL methods face significant challenges when training on static datasets~\cite{BCQ}. To address these overestimation issues, various offline RL techniques have been proposed~\cite{BCQ, IQL, TD3+BC, CQL}.
This study specifically evaluates the robustness of two common strategies for offline RL: policy constraints~\cite{BCQ,TD3+BC} and regularization~\cite{IQL}.

Batch-constrained Q-learning (BCQ)~\cite{BCQ} utilizes policy constraints to ensure the learned policy remains close to the behavior policy that generates the training data. 
Similarly, the Twin-Delayed DDPG with behavior cloning (TD3+BC)~\cite{TD3+BC}, an offline variant of TD3~\cite{TD3}, implicitly constrains the learned policy by adding a regularization term to the objective function. 
These policy constraints effectively limit the action space, preventing the agent from exploring unrealistic or unencountered actions in the dataset.
Implicit Q-learning (IQL)~\cite{IQL} regularizes the value function update process, relying on the state value distribution from the dataset rather than the learned policy.
This regularization mitigates the risk of overestimating Q-values, ensuring that the resulting policy remains within the empirical data distribution.
However, both policy constraints and regularization strategies may lose robustness against random and adversarial perturbations because the training data does not contain data adapted to such perturbations.

\subsection{Robust reinforcement learning}
\label{sec:2-2}
The robustness of RL, especially in robot control, mainly stems from addressing the reality gap. This gap arises from discrepancies in physical parameters such as friction coefficients, terrain geometry, and mechanical failures.
In the context of online RL, one approach to address the reality gap involves leveraging the Robust Markov Decision Process (RMDP) framework~\cite{RMDP_1,RMDP_2,RMDP_3,RMDP_4,RMDP_5}, which relies on a set of state transition models known as an uncertainty set. RMDP seeks to derive an optimal strategy by considering the worst-case scenarios within this uncertainty set. However, the high computational cost of RMDP makes it impractical for large state-action spaces.
A more practical approach to bridging this gap involves introducing perturbations during the training process.
Domain randomization~\cite{domain_random_1,domain_random_2,domain_random_3} generates diverse environments by randomizing parameters such as friction coefficients, thereby training the model to adapt to these changes.  Alternatively, adversarial training~\cite{adv_trng_1,adv_trng_2,adv_trng_3,adv_trng_4,adv_trng_5} improves robustness by introducing adversarial attacks that intentionally reduce rewards.
Differential evolution has also been employed to generate adversarial perturbations~\cite{o2022_adv}, leading to significant reward reductions in walking robot scenarios. Curriculum-based adversarial learning~\cite{adv_trng_2} enhances robustness by applying adversarial attacks to the action space.

In offline RL, robustness can be categorized into training and testing time robustness.
Training-time robustness addresses corrupted training data~\cite{robust_trngtime_1,robust_trngtime_2,robust_trngtime_3}.
For example, IQL is more robust than other offline RL methods for training-time robustness~\cite{robust_trngtime_3}.
Furthermore, a variant, Robust-IQL, incorporates normalization, Huber loss, and quantile Q estimator to enhance resistance against corrupted transitions.
Several studies have focused on the testing-time robustness in an environment with perturbations that are not represented in training datasets~\cite{robust_testtime_1,robust_testtime_2,robust_testtime_3,robust_testtime_4,RORL,MICRO,robust_testtime_5}. 
In relation to our work, Yan et al.~\cite{RORL} enhances robustness against adversarial attacks in the state space by smoothing the action value distribution and employing conservative value estimates. 
Liu et al.~\cite{MICRO} proposes a model-based approach to address adversarial state perturbations using ensemble learning with a set of state transition functions.
Nguyen et al.~\cite{robust_testtime_5} demonstrate the vulnerability of offline RL models to adversarial examples, where the outputs of critic and actor networks deviate from clean examples based on KL divergence. The study introduces a defense regularization term in the loss function to minimize the mean squared error between adversarial and clean example outputs, showing improved robustness against adversarial attacks.

Nevertheless, research on adversarial attacks specifically targeting action spaces and corresponding defense strategies
remains limited.
This study explores testing-time robustness against both adversarial and random perturbations in the context of model-free control for legged robots.

\subsection{Comparison between offline and online reinforcement learning}
\label{sec:2-3}
Offline RL and online RL represent fundamentally different learning scheme in how data is collected and utilized. Offline RL trains policies solely on fixed datasets, whereas online RL continuously gathers new experience through environment interaction and updates the policy accordingly~\cite{offline_rl_1,offline_rl_2}. These differences can lead to significant discrepancies in generalization, adaptability, and robustness.

Sujit et al.~\cite{Sujit2023SeqEval} point out that such discrepancies in evaluation protocols make direct comparisons between offline RL and online RL difficult, and propose a unified evaluation framework (SeqEval) based on data usage and interaction level. Mediratta et al.~\cite{Mediratta2024GenGap} demonstrate that even when trained on expert-level datasets, offline RL often underperforms in unseen environments due to limited data coverage and the absence of exploration.
Brandi et al.~\cite{Brandi2022MPC} provide a practical comparison between the two paradigms in a real-world control task. Their results show that offline RL benefits from stable early performance due to model pretraining, while online RL, though initially less effective, eventually achieves comparable performance through continuous interaction. This reveals a trade-off between initial stability and long-term adaptability.
Although not directly compared to the online RL, further studies have examined structural limitations of offline RL. Schweighofer et al.~\cite{Schweighofer2022DatasetQuality} show that dataset properties such as trajectory quality and state-action coverage significantly affect offline RL performance, highlighting its dependence on the training data. In contrast, online RL inherently collects diverse data via exploration. Park et al.~\cite{Park2024ValueLearningBottleneck} argue that the main bottleneck in offline RL lies not in value estimation, but in the limited generalization capacity of the learned policy, particularly in unseen states during deployment.

These studies collectively highlight the fundamental differences between offline and online RL. Building on these insights, this study empirically investigates their testing-time robustness under random and adversarial perturbations in continuous control tasks.

\section{Method}
\label{sec:3}
\begin{figure*}[t]
    \centering
    \includegraphics[width=1.0\hsize]{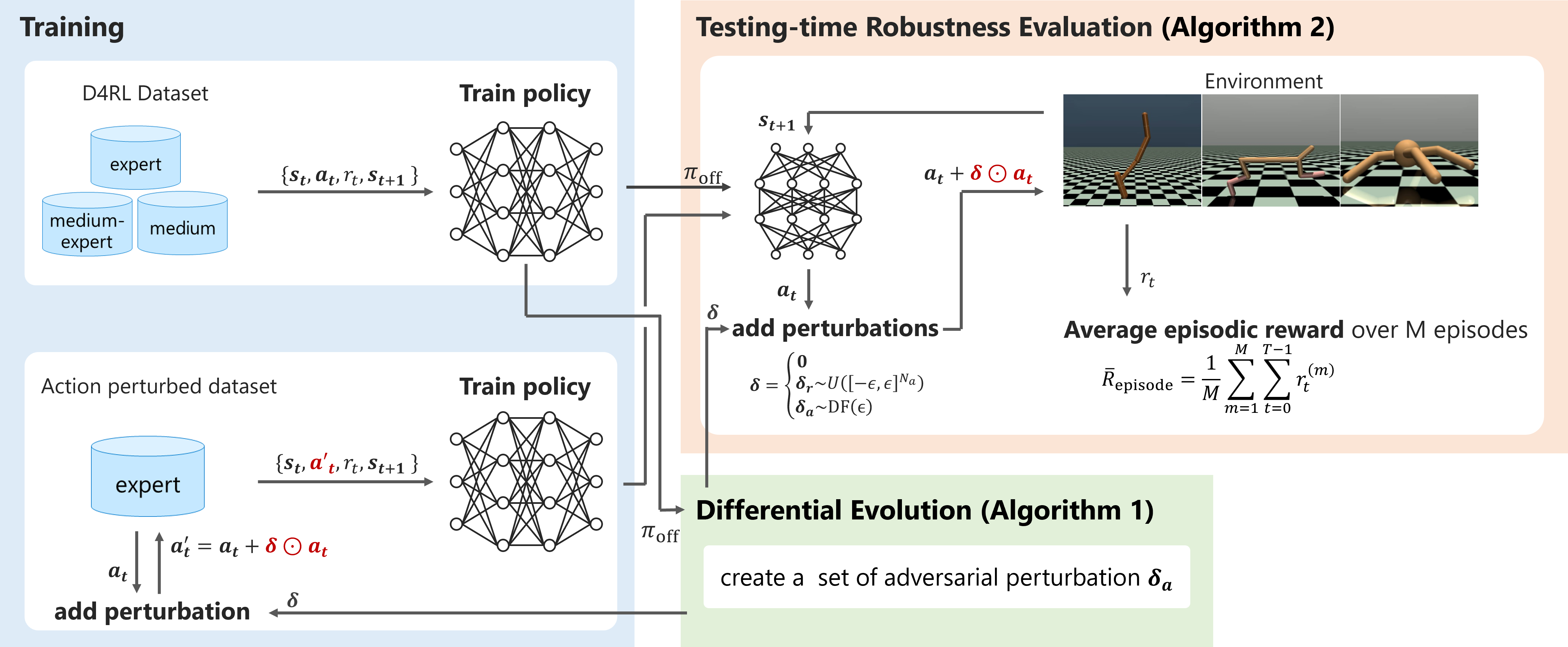}
    \caption{Overview of the robustness evaluation for offline RL. Offline RL models are trained on varying-quality offline datasets and evaluated on forward walking tasks in diverse robot environments. Testing-time robustness is evaluated by average episodic rewards under three conditions: normal (no perturbations), random, or adversarial perturbations applied to actions. Adversarial perturbations, generated using the differential evolution algorithm, minimize the average episodic reward over multiple generations in the same robot environment. These adversarial perturbations are also used to create action-perturbed datasets, which are tested to determine whether training with such datasets improves testing-time robustness.}
    \label{fig:fig1_robust_eval_overview}
\end{figure*}
We evaluate the testing-time robustness of offline RL methods by conducting forward walking tasks with legged robots in three distinct environments, as shown in Figure~\ref{fig:fig1_robust_eval_overview}. These environments include three perturbation scenarios: normal (no perturbations), random perturbations, and adversarial perturbations, both applied to the robot's joint torque signals. For the robustness evaluation, we employ two common approaches for offline RL: policy constraints and regularization.

\subsection{Offline reinforcement learning}
\label{sec:3-1}
Reinforcement learning is modeled as a Markov decision process (MDP). An MDP is defined by a tuple $(\mathcal{S},\mathcal{A}, P,{P_0}, r,\gamma)$ where $\mathcal{S}$ is a set of states, $\mathcal{A}$ is a set of actions, $P:\mathcal{S} \times \mathcal{A} \times \mathcal{S} \rightarrow [0, 1]$ defines the transition probability $P(\vect{s}'|\vect{s}, \vect{a})$ from state $\vect{s}$ to state $\vect{s}'$ given action $\vect{a}$, $P_0:\mathcal{S}\rightarrow [0,1]$ denotes the initial state distribution, $r:\mathcal{S} \times \mathcal{A} \rightarrow \mathbb{R}$ is the reward function that assigns the immediate reward $r(\vect{s}, \vect{a})$ received after transitioning, and $\gamma \in [0, 1]$ is the discount factor that determines the importance of future rewards.
The goal of an MDP is to find an optimal policy $\pi^*$ that maximizes the expected sum of rewards.

In the offline RL setting, we are provided with dataset ${\mathcal{D}}=\{(\vect{s}_t,\vect{a}_t,\vect{s}_{t+1},r_t)_i\}$, which may be sourced from previous learning episodes, expert demonstrations, or even random actions.
Here, the subscript $t$ denotes a timestep within an episode, and the subscript $i$ indexes individual transitions in the dataset.
The dataset ${\mathcal{D}}$ is assumed to be generated according to a behavior policy $\pi_\beta$. The objective of offline RL is to find an optimal policy $\pi^*$ using only the dataset ${\mathcal{D}}$ without further environmental interactions.
A principal challenge in offline RL is the distributional shift problem. This problem arises when the dataset ${\mathcal{D}}$ does not accurately reflect real-world situations or different scenarios. Such static datasets may result in misguided learning, leading to policies that perform poorly in new or unseen environments. 
To mitigate the effects of the distribution shift, several methods have been proposed\cite{BCQ,IQL,TD3+BC,CQL}.

Policy constraint methods impose constraints to minimize the discrepancy between the learned policy $\pi_\text{off}$ and the
behavior policy $\pi_\beta$.
Similarly, policy regularization methods penalize deviations from behavior policy $\pi_\beta$ by incorporating regularization terms into the objective function.
These constraints and regularization terms help to keep the learned policy within the support of the behavior policy, and ensure that the learned policy does not make decisions that deviate significantly from dataset $\mathcal{D}$.  
However, this conservative strategy may backfire, potentially leading to vulnerabilities in offline RL models during testing phases.

\subsection{Action perturbation}
\label{sec:3-2}
For action attack, a perturbation $\vect{\delta}$ is applied to the action $\vect{a}_t$ as follows: 
\begin{equation}
    \label{equ:attack}
    \vect{a}_{t}'=\vect{a}_{t}+\vect{\delta}\odot\vect{a}_{t},
\end{equation}
where $t$ denotes the time step in an episode, and $\odot$ denotes the Hadamard product, which represents the element-wise product of two vectors.

In this study, we consider three conditions: normal (no perturbations), uniform random perturbations, and adversarial perturbations.
These perturbations are generated as follows:
\begin{equation}
    \label{equ:perturbations_case}
    \vect{\delta} = 
        \begin{cases}
            \vect{0} & \text{normal},\\
            \vect{\delta}_r\sim U([-\epsilon,\epsilon]^{N_a}) & \text{random perturbations,}\\
            \vect{\delta}_a\sim \text{DE}(\epsilon) & \text{adversarial perturbations}.
        \end{cases}
\end{equation}
In the normal condition, the perturbation $\vect{\delta}$ is set to the zero vector $\vect{0}$, meaning no perturbations are applied to the actions $\vect{a}$.
The random perturbation $\vect{\delta}_r$ is drawn from the uniform distribution $U([-\epsilon,\epsilon]^{N_a})$, where $N_a$ denotes the dimension of $\vect{a}$, corresponding to the number of actuators, and $\epsilon$ specifies the perturbation strength. See Appendix A for details of the perturbation strength configuration. Each element of $\vect{a}$ is independently perturbed by a random number drawn from the uniform distribution over the interval $[-\epsilon,\epsilon]$.
As illustrated in the bottom-right corner of Figure~\ref{fig:fig1_robust_eval_overview}, the adversarial perturbation $\vect{\delta}_a$ is obtained using the differential
evolution (DE) method~\cite{o2022_adv}, denoted by $\text{DE}(\epsilon)$.

\begin{algorithm}[t]
    \caption{Differential Evolution for Adversarial Attack}
    \label{algo:DE}
    \begin{algorithmic}[1]
        \REQUIRE $NP$: Population size
        \REQUIRE $M$: Maximum number of episodes 
        \STATE Generate initial population $\{\vect{\delta}_0^{(1)}, \vect{\delta}_0^{(2)}, \dots, \vect{\delta}_0^{(NP)}\}$ with $\vect{\delta}_0^{(i)}\sim U\left([-\epsilon,\epsilon]^{N_a}\right)$
        \STATE $R_{\text{min}}\leftarrow\infty$
        \FOR {each generation $g = 1,2,\ldots$}
            \FOR {each individual $i = 1,2,\ldots, NP$}
                \STATE Generate trial individual $\vect{u}_g^{(i)}$ by mutation and crossover using Equation~(\ref{equ:mutation}) and Equation~(\ref{equ:crossover})
                \STATE $\vect{u}_g^{(i)}\leftarrow\text{clip}_{[-\epsilon,\epsilon]}(\vect{u}_g^{(i)})$
                \FOR{each episode $m=1,2,\ldots, M$}
                    \STATE Compute the average episodic reward of the g-th generation $\bar{R}_g$ and target $\bar{R}_{g-1}$ using Equation~(\ref{equ:average_epsidic_reward})
                \ENDFOR
                \IF {$\bar{R}_g\leq\bar{R}_{g-1}$}
                    \STATE Accept trial individual $\vect{\delta}_g^{(i)}\leftarrow\vect{u}_g^{(i)}$
                    \IF{$\bar{R}_g\leq R_{\text{min}}$}
                        \STATE $R_{\text{min}}\leftarrow \bar{R}_g$
                        \STATE Update best individual $\vect{\delta}_{\text{best}}\leftarrow\vect{\delta}_g^{(i)}$
                    \ENDIF
                \ELSE
                    \STATE Retain target individual $\vect{\delta}_g^{(i)}\leftarrow\vect{\delta}_{g-1}^{(i)}$
                \ENDIF
            \ENDFOR
        \ENDFOR
    \end{algorithmic}
\end{algorithm}

The DE method is represented as Algorithm~\ref{algo:DE}, with perturbations for the $g$-th generation denoted by $\{\vect{\delta}_g^{(1)}, \vect{\delta}_g^{(2)}, \ldots, \vect{\delta}_g^{(NP)}\}$,
where $NP$ denotes the number of individuals for each generation.
Each perturbation $\vect{\delta}_g^{(i)}$ is evaluated based on the average episodic reward $\bar{R}_{\text{episode}}$ computed over $M$ episodes as:
\begin{align}
    \label{equ:average_epsidic_reward}
        \bar{R}_{\text{episode}} = \frac{1}{M}\sum^{M}_{m=1}\sum^{T-1}_{t=0} r_t^{(m)},
\end{align}
where $r_t^{(m)}=r(\vect{s}_{t},\vect{a}_{t}')$ denotes the immediate reward at time $t$ for episode $m$.
We conduct over 100 trials for each generation under the trained policy $\pi_\text{off}$.
The DE algorithm uses mutation and crossover processes to generate new perturbations.
For mutation, the $i$-th mutant individual $\vect{v}_{g+1}^{(i)}$ is calculated as:
\begin{align}
    \label{equ:mutation}
    \vect{v}_{g+1}^{(i)} = 
        \vect{\delta}_\text{best}
        + F(\vect{\delta}_g^{(r_1)}-\vect{\delta}_g^{(r_2)}), 
\end{align}
where $F\in(0.5,1]$ is a random scaling factor, $r_1$ and $r_2$ are mutually exclusive indices randomly chosen from $\{1,2,\ldots,NP\}$.
For crossover, the binominal crossover is used to compute the $j$-th element of the $i$-th trial individual as follows: 
\begin{equation}
    \label{equ:crossover}
    u_{j,g+1}^{i} = 
        \begin{cases}
            v_{j,g+1}^{i} & \text{if $r\leq CR$ or $j=j_r$},\\
            \delta_{j,g}^{i} & \text{otherwise},
        \end{cases}
\end{equation}
where $r\in[0,1]$ is a uniform random number, $CR=0.7$ is the crossover constant, and $j_r$ is an index randomly chosen from $\{1,2,\ldots,N_a\}$.
Each element of $\vect{u}_{g+1}^{(i)}$ is constrained to the range of $[-\epsilon,\epsilon]$ using the clip function, defined by $\text{clip}_{[-\epsilon,\epsilon]}(\vect{x})=\max{(\min{(\vect{x},\epsilon)},-\epsilon)}$.
Finally,  the individual from the final generation is utilized as the adversarial perturbation $\vect{\delta}_a$.

\begin{algorithm}[t]
    \caption{Testing-Time Robustness Evaluation}
    \label{algo:Robust_eval}
    \begin{algorithmic}[1]
        \REQUIRE $\pi_{\mathrm{off}}$: Trained policy with offline RL method
        \REQUIRE $M$: Maximum number of episodes
        \FOR {each episode $m = 1,2,\ldots,M$}
            \STATE Generate perturbation $\vect{\delta}$ using Equation~(\ref{equ:perturbations_case})
            \STATE Sample initial state $\vect{s}_0\sim P_0(\cdot)$
            \STATE Set $t\leftarrow0$ and $R^{(m)} \leftarrow0$
            \WHILE{not terminated}
                \STATE Sample action $\vect{a}_t\sim\pi_{\mathrm{off}}(\cdot|\vect{s}_t)$
                \STATE Obtain reward $r_t\leftarrow r(\vect{s}_t,\vect{a}_t+\vect{\delta}\odot\vect{a}_t)$ 
                \STATE Accumulate reward $R^{(m)}\leftarrow R^{(m)}+r_t$
                \STATE
                Sample state $\vect{s}_{t+1}\sim P(\cdot|\vect{s}_{t},\vect{a}_t)$
                \STATE $t\leftarrow t+1$
            \ENDWHILE
        \ENDFOR
        \RETURN $\bar{R}_{\text{episode}}=\frac{1}{M}\sum^{M}_{m=1}R^{(m)}$
        \end{algorithmic}
\end{algorithm}

\subsection{Testing-time robustness evaluation}
\label{sec:3-3}

The procedure for evaluating the testing-time robustness of offline RL methods is outlined in Algorithm~\ref{algo:Robust_eval} and top-right corner of Figure~\ref{fig:fig1_robust_eval_overview}. 
The evaluation is conducted for the trained policy $\pi_\text{off}$ under three  perturbation conditions in Equation~(\ref{equ:perturbations_case}). 
We assume that perturbations $\vect{\delta}$ remain constant throughout each episode;  $\vect{\delta}$ is initialized at the beginning of the episode and remains unchanged until the end. 
The robustness is evaluated using the average episodic reward over $M$ episodes in Equation~(\ref{equ:average_epsidic_reward}).
Each episode terminates when either the time steps $T$ reaches $1000$ or the robot falls over.

\section{Experiments}
\label{sec:4}
This section evaluates the testing-time robustness of offline RL methods through three main experiments. First, we compare the testing-time robustness of offline RL methods to online RL methods by measuring the average episodic rewards over 1,000 trials under three distinct conditions: the normal condition (no perturbations), the random condition (uniform random perturbations), and the adversarial condition (adversarial perturbations). 
Second, we examine how state-action coverage in training datasets impacts the testing-time robustness of offline RL methods. 
Finally, we investigate whether training policies on action-perturbed datasets enhance their robustness against the perturbations.

\subsection{Experimental setup}
\label{sec:4-1}
\paragraph{RL methods}: We employ existing offline RL methods (BCQ, TD3+BC, IQL) implemented in the Data-Driven Deep Reinforcement Learning library for Python (d3rlpy)~\cite{d3rlpy}, a framework designed for offline RL.
Additionally, we employ three widely-used online RL methods: Proximal Policy Optimization (PPO)~\cite{PPO}, Soft Actor-Critic (SAC)~\cite{SAC}, and TD3~\cite{TD3}. 
The hyperparameter configurations are adapted from the original paper~\cite{PPO} and the prior study~\cite{ppo_param}
For SAC and TD3, we leverage the implementation provided by the d3rlpy library.
See Appendix B for details of hyperparameter configurations.

\paragraph{Training datasets}: 
Training is conducted using datasets from Datasets for Deep Data-Driven Reinforcement Learning (D4RL)~\cite{D4RL}, which include three quality levels: expert, medium-expert, and medium.
The expert and medium datasets consist of 1 million trajectories collected using well-trained and less-trained policies, respectively.
The medium-expert dataset combines the expert and medium datasets, resulting in a total of 2 million trajectories.
We create two types of action-perturbed datasets for all legged robot environments: randomly-perturbed datasets and adversarially-perturbed datasets. Random and adversarial perturbations, as defined in Equation~(\ref{equ:perturbations_case}), are applied to all action data within the D4RL expert dataset, following Equation~(\ref{equ:attack}). 
See Appendix C for histograms of action distributions in each action-perturbed dataset.

\paragraph{Training steps}:
For the expert dataset, RL models are trained for 3 million steps using BCQ and IQL methods, respectively, whereas models using TD3+BC require extended training of up to 5 million steps to ensure convergence.
With the medium-expert dataset, all RL models are trained for 3 million steps.
For the medium dataset, RL models are trained for 3 million steps using BCQ and IQL, whereas models using TD3+BC require only up to 2 million steps of training.
For the action-perturbed datasets, RL models using each offline RL method were trained for the same number of steps as those trained on the expert dataset.
Regarding the online RL methods, RL models using PPO are trained as described in ~\cite{o2022_adv} and ~\cite{ppo_param}.
RL models are trained for over 3 million steps using SAC and TD3 methods.

\begin{figure}[t]
    \centering
     \includegraphics[width=1.0\hsize]{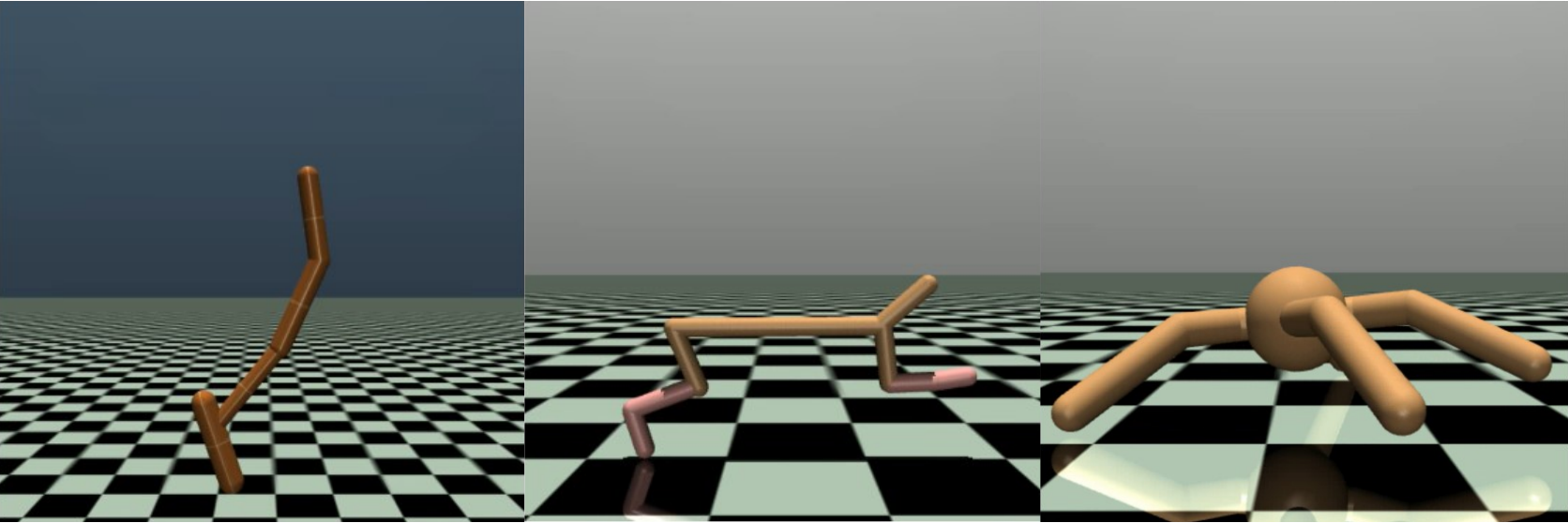}
    \caption{Legged robot environments: Hopper-v2 (left), HalfCheetah-v2 (center), and Ant-v2 (right).}
    \label{fig:fig2_legged_robot_envs}
\end{figure}

\paragraph{Legged robot environments}:
Forward walking tasks are conducted in the Hopper-v2, HalfCheetah-v2, and Ant-v2 environments of OpenAI Gym~\cite{OpenAI}, which serve as widely recognized as benchmark environments in offline RL~\cite{BCQ,IQL,TD3+BC,CQL}. These tasks are simulated within the MuJoCo physics engine, as illustrated in Figure~\ref{fig:fig2_legged_robot_envs}.

The reward functions for each robot are defined as follows: 
\begin{align}
    \label{equ:hop_reward}
    r(\vect{s},\vect{a})=
    \begin{cases}
    v_{\text{fwd}}-0.001\left\|\vect{a}\right\|^{2}+1,\ \text{Hopper}\\
    v_{\text{fwd}}-0.1\left\|\vect{a}\right\|^{2}+1,\  \text{HalfCheetah}\\
    v_{\text{fwd}}-0.5\|\vect{a}\|^{2}-0.5\cdot10^{-3}\left\|\vect{f}\right\|^{2}+1,\  \text{Ant}
    \end{cases}
\end{align}
where $v_{\text {fwd }}$ is the forward walking speed and $\vect{f}$ denotes the contact force. 

\paragraph{Perturbation setting}:
Perturbations under each condition are defined according to Equation~(\ref{equ:perturbations_case}).
In the Hopper-v2 and HalfCheetah-v2 environments, the strengths of $\vect{\delta}_r$ and $\vect{\delta}_a$ are set to $\epsilon = 0.3$ for all experiments.
In the Ant-v2 environment, the perturbation strength is set to $\epsilon = 0.5$ for most experiments. However, for the final experiment involving the action-perturbed dataset, the perturbation strength is reduced to $\epsilon=0.3$ due to insufficient training with the strength $\epsilon=0.5$.
The adversarial perturbation $\vect{\delta}_a$ in Equation~(\ref{equ:perturbations_case}) is generated using the DE method over 30 generations. The population sizes for DE are set to 45, 90, and 120 for Hopper-v2, HalfCheetah-v2, and Ant-v2, respectively.

\begin{table}[t]
    \caption{Testing-time robustness evaluation results of offline RL (BCQ, IQL, TD3+BC) and online RL (PPO, SAC, TD3). The evaluation is conducted in three legged robot environment (Hopper-v2, HalfCheetah-v2, Ant-v2) under normal, random, and adversarial conditions. Offline RL methods are trained with the expert dataset. Results are presented as average episodic rewards $\pm$ standard deviations. The values reported in the average row represent average results of three offline RL methods under each experiment setting.}
    \centering
    {\begin{tabular}{lllrrr}
        \toprule            
        \multirow{2}{*}{Environment}&\multirow{2}{*}{Training datasets}&\multirow{2}{*}{RL methods}&\multicolumn{3}{c}{Perturbation conditions}\\
        \cmidrule(lr){4-6}
        &&&\multicolumn{1}{c}{Normal}&\multicolumn{1}{c}{Random}&\multicolumn{1}{c}{Adversarial} \\
        \midrule
        \multirow{8}{*}{Hopper-v2}&\multirow{4}{*}{expert}
            &BCQ&$2371\pm449$&$1556\pm1296$&$282\pm4$ \\ 
            &&TD3+BC&$3597\pm93$&$2027\pm1394$&$424\pm78$ \\
            &&IQL&$1631\pm597$&$1290\pm939$&$421\pm24$ \\
            \cline{3-6}
            &&Average&$2533$&$1624$&$376$ \\ 
        \cline{2-6}
        &\multirow{4}{*}{online}
            &PPO&$2529\pm109$&$1456\pm935$&$346\pm6$ \\
            &&SAC&$3148\pm746$&$1945\pm1294$&$445\pm89$ \\
            &&TD3&$3439\pm2$&$2706\pm1130$&$487\pm18$ \\
            \cline{3-6}
            &&Average&$3039$&$2036$&$426$ \\ 
        \midrule
        \multirow{8}{*}{HalfCheetah-v2}&\multirow{4}{*}{expert}
            &BCQ&$11414\pm1406$&$8036\pm3179$&$244\pm262$ \\ 
            &&TD3+BC&$11713\pm421$&$9034\pm2337$&$1227\pm999$ \\
            &&IQL&$9316\pm2214$&$6198\pm3057$&$313\pm366$ \\
            \cline{3-6}
            &&Average&$10814$&$7756$&$595$ \\ 
        \cline{2-6}
        &\multirow{4}{*}{online}
            &PPO&$6822\pm32$&$6579\pm161$&$5985\pm107$ \\
            &&SAC&$7664\pm58$&$5610\pm1069$&$2841\pm71$ \\
            &&TD3&$7076\pm289$&$5801\pm646$&$3713\pm106$ \\
            \cline{3-6}
            &&Average&$7187$&$5997$&$4180$ \\ 
        \midrule
        \multirow{8}{*}{Ant-v2}&\multirow{4}{*}{expert}
            &BCQ&$4511\pm1568$&$3102\pm1781$&$-1129\pm1083$ \\
            &&TD3+BC&$4715\pm1676$&$3317\pm1822$&$-1209\pm1089$ \\
            &&IQL&$4175\pm1546$&$2851\pm1731$&$-1503\pm1026$ \\
            \cline{3-6}
            &&Average&$4467$&$3090$&$-1280$ \\ 
        \cline{2-6}
        &\multirow{4}{*}{online}
            &PPO&$5698\pm989$&$4975\pm1191$&$2654\pm1748$ \\
            &&SAC&$6070\pm890$&$4226\pm1611$&$942\pm888$ \\
            &&TD3&$5154\pm1100$&$3349\pm1464$&$685\pm1116$ \\
            \cline{3-6}
            &&Average&$5641$&$4183$&$1427$ \\ 
        \bottomrule
    \end{tabular}}
    \label{tab:tab1_resuls_exp_vs_online}
\end{table}

\begin{table}[t]
    \caption{Testing-time robustness evaluation results of offline RL trained on the medium-expert and the medium datasets. The evaluation is conducted in three legged robot environments (Hopper-v2, HalfCheetah-v2, Ant-v2) under normal, random, and adversarial conditions. Results are presented as the average episodic rewards $\pm$ standard deviations. The values reported in the average row represents average results of three offline RL methods under each experiment setting.}
    \centering
    {\begin{tabular}{lllrrr}
        \toprule            
        \multirow{2}{*}{Environment}&\multirow{2}{*}{Training datasets}&\multirow{2}{*}{RL methods}&\multicolumn{3}{c}{Perturbation conditions}\\
        \cmidrule(lr){4-6}
        &&&\multicolumn{1}{c}{Normal}&\multicolumn{1}{c}{Random}&\multicolumn{1}{c}{Adversarial} \\
        \midrule
        \multirow{12}{*}{Hopper-v2}&\multirow{4}{*}{expert}
            &BCQ&$2371\pm449$&$1556\pm1296$&$282\pm4$ \\ 
            &&TD3+BC&$3597\pm93$&$2027\pm1394$&$424\pm78$\\
            &&IQL&$1631\pm597$&$1290\pm939$&$421\pm24$\\
            \cline{3-6}
            &&Average&$2533$&$1624$&$376$ \\ 
        \cline{2-6}
        &\multirow{4}{*}{medium-expert}
            &BCQ&$2928\pm973$&$1581\pm1063$&$453\pm57$ \\
            &&TD3+BC&$3059\pm867$&$1706\pm1151$&$500\pm57$ \\
            &&IQL&$879\pm277$&$790\pm454$&$432\pm44$ \\
            \cline{3-6}
            &&Average&$2289$&$1262$&$700$ \\
        \cline{2-6}
        &\multirow{4}{*}{medium}
            &BCQ&$2030\pm502$&$1483\pm580$&$609\pm12$  \\
            &&TD3+BC&$1969\pm398$&$1569\pm560$&$741\pm14$ \\
            &&IQL&$1200\pm205$&$1116\pm439$&$598\pm28$ \\
            \cline{3-6}
            &&Average&$1733$&$1389$&$649$ \\
        \midrule
        \multirow{12}{*}{HalfCheetah-v2}&\multirow{4}{*}{expert}
            &BCQ&$11414\pm1406$&$8036\pm3179$&$244\pm262$ \\
            &&TD3+BC&$11713\pm421$&$9034\pm2337$&$1227\pm999$ \\
            &&IQL&$9316\pm2214$&$6198\pm3057$&$313\pm366$ \\
            \cline{3-6}
            &&Average&$10814$&$7756$&$595$ \\
        \cline{2-6}
        &\multirow{4}{*}{medium-expert}
            &BCQ&$10566\pm1563$&$7139\pm2448$&$1135\pm948$ \\
            &&TD3+BC&$11729\pm724$&$9494\pm1994$&$2273\pm1470$ \\
            &&IQL&$6974\pm1780$&$5683\pm1583$&$2044\pm382$ \\
            \cline{3-6}
            &&Average&$9756$&$7439$&$1817$ \\
        \cline{2-6}
        &\multirow{4}{*}{medium}
            &BCQ&$5504\pm324$&$5077\pm338$&$3462\pm405$ \\
            &&TD3+BC&$5739\pm81$&$5329\pm328$&$3568\pm135$ \\
            &&IQL&$5364\pm267$&$4903\pm464$&$2814\pm240$ \\
            \cline{3-6}
            &&Average&$5536$&$5103$&$3281$ \\
        \midrule
        \multirow{12}{*}{Ant-v2}&\multirow{4}{*}{expert}
            &BCQ&$4511\pm1568$&$3102\pm1781$&$-1129\pm1083$ \\
            &&TD3+BC&$4715\pm1676$&$3317\pm1822$&$-1209\pm1089$ \\
            &&IQL&$4175\pm1546$&$2851\pm1731$&$-1503\pm1026$ \\
            \cline{3-6}
            &&Average&$4467$&$3090$&$-1280$ \\
        \cline{2-6}
        &\multirow{4}{*}{medium-expert}
            &BCQ&$4250\pm1568$&$3397\pm1343$&$1467\pm1571$ \\
            &&TD3+BC&$4588\pm1718$&$3818\pm1431$&$1837\pm1505$ \\
            &&IQL&$4131\pm1457$&$3235\pm1352$&$1074\pm605$ \\
            \cline{3-6}
            &&Average&$4323$&$3483$&$1459$ \\
        \cline{2-6}
        &\multirow{4}{*}{medium}
            &BCQ&$4219\pm709$&$3361\pm1045$&$2597\pm376$ \\
            &&TD3+BC&$4194\pm1294$&$3433\pm1340$&$1206\pm998$ \\
            &&IQL&$3527\pm997$&$2946\pm979$&$1329\pm992$ \\
            \cline{3-6}
            &&Average&$3980$&$3247$&$1711$ \\
        \bottomrule
    \end{tabular}}
    \label{tab:tab2_results_coverage_ds}
\end{table}

\begin{table}[t]
    \caption{Testing-time robustness evaluation results of offline RL trained on action-perturbed datasets. The evaluation is conducted in three legged robot environments (Hopper-v2, HalfCheetah-v2, Ant-v2) under normal, random, and adversarial conditions. Results are presented as average episodic rewards $\pm$ standard deviations. The values reported in the average row represents average results of three offline RL methods under each experiment setting. \textit{randomly-perturbed} refers to 
    the expert dataset with randomly-perturbed actions, and \textit{adversarially-perturbed} refers to the expert dataset with adversarially-perturbed actions.}
    \centering
    {\begin{tabular}{lllrrr}
        \toprule            
        \multirow{2}{*}{Environment}&\multirow{2}{*}{Training datasets}&\multirow{2}{*}{RL methods}&\multicolumn{3}{c}{Perturbation conditions}\\
        \cmidrule(lr){4-6}
        &&&\multicolumn{1}{c}{Normal}&\multicolumn{1}{c}{Random}&\multicolumn{1}{c}{Adversarial} \\
        \midrule
        \multirow{12}{*}{Hopper-v2}&\multirow{4}{*}{expert}
            &BCQ&$2371\pm449$&$1556\pm1296$&$282\pm4$ \\
            &&TD3+BC&$3597\pm93$&$2027\pm1394$&$424\pm78$ \\
            &&IQL&$1631\pm597$&$1290\pm939$&$421\pm24$ \\
            \cline{3-6}
            &&Average&$2533$&$1624$&$376$ \\
        \cline{2-6}
        &\multirow{4}{*}{randomly-perturbed}
            &BCQ&$2858\pm717$&$1774\pm1232$&$506\pm82$ \\
            &&TD3+BC&$3594\pm84$&$2077\pm1350$&$413\pm13$ \\
            &&IQL&$1124\pm335$&$1006\pm642$&$413\pm34$ \\
            \cline{3-6}
            &&Average&$2525$&$1619$&$444$ \\
        \cline{2-6}
        &\multirow{4}{*}{adversarially-perturbed}
            &BCQ&$534\pm25$&$659\pm355$&$329\pm18$ \\
            &&TD3+BC&$353\pm5$&$411\pm195$&$336\pm3$ \\
            &&IQL&$474\pm54$&$498\pm182$&$333\pm38$ \\
            \cline{3-6}
            &&Average&$454$&$523$&$333$ \\
        \midrule
        \multirow{12}{*}{HalfCheetah-v2}&\multirow{4}{*}{expert}
            &BCQ&$11414\pm1406$&$8036\pm3179$&$244\pm262$ \\
            &&TD3+BC&$11713\pm421$&$9034\pm2337$&$1227\pm999$ \\
            &&IQL&$9316\pm2214$&$6198\pm3057$&$313\pm366$ \\
            \cline{3-6}
            &&Average&$10814$&$7756$&$595$ \\
        \cline{2-6}
        &\multirow{4}{*}{randomly-perturbed}
            &BCQ&$11110\pm1411$&$7660\pm3167$&$182\pm315$ \\
            &&TD3+BC&$11334\pm111$&$8753\pm2202$&$828\pm515$ \\
            &&IQL&$3618\pm2611$&$2367\pm2102$&$302\pm377$ \\
            \cline{3-6}
            &&Average&$8687$&$6260$&$437$ \\
        \cline{2-6}
        &\multirow{4}{*}{adversarially-perturbed}
            &BCQ&$427\pm648$&$1713\pm2839$&$43\pm54$ \\
            &&TD3+BC&$1737\pm1804$&$2840\pm3445$&$-26\pm68$ \\
            &&IQL&$407\pm445$&$863\pm1242$&$45\pm193$ \\
            \cline{3-6}
            &&Average&$857$&$1805$&$21$ \\
        \midrule
        \multirow{12}{*}{Ant-v2}&\multirow{4}{*}{expert}
            &BCQ&$4511\pm1568$&$4127\pm1536$&$1670\pm1769$ \\
            &&TD3+BC&$4715\pm1676$&$4189\pm1745$&$1432\pm1484$ \\
            &&IQL&$4175\pm1546$&$3707\pm1535$&$1966\pm1823$ \\
            \cline{3-6}
            &&Average&$4467$&$4008$&$1689$ \\
        \cline{2-6}
        &\multirow{4}{*}{randomly-perturbed}
            &BCQ&$4662\pm1486$&$4181\pm1467$&$2208\pm1863$ \\
            &&TD3+BC&$5266\pm1176$&$4534\pm1442$&$2292\pm1898$ \\
            &&IQL&$3385\pm1759$&$3231\pm1634$&$1609\pm1652$ \\
            \cline{3-6}
            &&Average&$4438$&$3982$&$2036$ \\
        \cline{2-6}
        &\multirow{4}{*}{adversarially-perturbed}
            &BCQ&$1609\pm1685$&$2351\pm1956$&$621\pm1530$ \\
            &&TD3+BC&$-159\pm534$&$-209\pm609$&$-428\pm815$ \\
            &&IQL&$1758\pm1628$&$2236\pm1831$&$505\pm1414$ \\
            \cline{3-6}
            &&Average&$1069$&$1459$&$233$ \\
        \bottomrule
    \end{tabular}}
    \label{tab:tab3_results_perturbed_ds}
\end{table}

\subsection{Comparing testing-time robustness: offline RL vs.~online RL}
\label{sec:4-2}
We compare the testing-time robustness of offline RL to online RL. 
Offline RL models are generally expected to be more vulnerable than online RL models because online RL models are typically trained on diverse state-action pairs through direct interaction with the environment. 
As shown below, the experimental results support this expectation.
Table~\ref{tab:tab1_resuls_exp_vs_online} summarizes the experiment results for the Hopper-v2, HalfCheetah-v2, and Ant-v2 environments.
The table presents the average episodic rewards over 1,000 trials, along with the corresponding standard deviations. 

From Table~\ref{tab:tab1_resuls_exp_vs_online}, both offline RL (BCQ, TD3+BC, IQL) and online RL (PPO, SAC, TD3) exhibit lower average episodic rewards under the random perturbation condition compared to the normal condition, with further deterioration observed under the adversarial perturbation condition. 
This result demonstrates that the applied perturbations, particularly the adversarial ones, affect the legged robot control using both online and offline RL as expected.
In the Hopper-v2 environment, offline RL performs on par with online RL under both the normal and random conditions, but shows slightly lower performance under the adversarial condition. In contrast, in the HalfCheetah-v2 and Ant-v2 environments, offline RL yields results that are either comparable to or better than those of online RL under normal and random conditions. However, under adversarial conditions, the performance of offline RL significantly decreases, particularly in the Ant-v2 environment, where the average episodic reward falls below zero.

This finding suggests that offline RL brings more vulnerability due to insufficient experiential learning.
The significant reduction in average episodic rewards 
occurs consistently across different legged robot scenarios and the offline RL methods, emphasizing the challenges for robust offline RL.
The results further suggest that the vulnerability of offline RL stems primarily from the offline setting itself rather than from policy characteristics, as the offline RL methods exhibit greater vulnerability to adversarial perturbations than the online RL methods, regardless of whether the policies are stochastic and deterministic policies.

The vulnerabilities identified in this study highlight fundamental challenges intrinsic to offline reinforcement learning, rather than limitations tied to specific robot models or the MuJoCo environment.
These challenges stem from structural constraints of offline RL—particularly, its inability to incorporate feedback from action outcomes—which are independent of simulation platforms or actuation mechanisms.
For instance, in real-world legged robots, actuator lag, a delay between the control command and the actual torque output, is a common phenomenon. 
Such lag can cause the executed action to deviate from the intended action even if the control policy itself is correct. 
Since offline RL cannot observe the post-action distortion induced by actuator lag during training, the learned policy may fail catastrophically when deployed on hardware.
Therefore, in offline settings that rely on static training data, similar vulnerabilities may manifest in other robotic systems and real-world applications where interactive learning during training is not possible.

\subsection{Impact of state-action coverage on offline RL robustness}
\label{sec4-4}
Based on the robustness differences between offline RL and online RL identified in the previous section, we hypothesize that these differences stem from the dependency of offline RL training on the training dataset. 
Specifically, we hypothesize that variations in state-action coverage across different training datasets contribute to these robustness disparities. 
To investigate this hypothesis, we evaluate the robustness of offline RL methods trained on the D4RL medium-expert and medium datasets. Table~\ref{tab:tab2_results_coverage_ds} summarizes the results.

For the medium-expert dataset, average episodic rewards under the normal and random conditions are consistent with those of the expert dataset. However, under the adversarial condition, 
the medium-expert dataset yields higher average rewards and small standard deviations than the expert dataset.
This result indicates a shift toward a more positive reward distribution.

For the medium dataset, average episodic rewards in the HalfCheetah-v2 environment under the normal condition are approximately half of those for the expert dataset. In contrast, the hopper-v2 and Ant-v2 environments show only slight decreases compared to the expert dataset results.
This finding aligns with previous studies~\cite{CQL,TD3+BC,Fisher-BRC,D4RL}, which report reduced rewards for policies trained on the medium dataset in HalfCheetah-v2 environment.
This suggests that these differences reflect the specific characteristics of each legged robot environment.
Furthermore, under the adversarial condition, 
both the medium and medium-expert datasets yield significant increases in average rewards compared to the expert dataset. 
Moreover, in many cases, the medium dataset even outperforms the medium-expert dataset in terms of average rewards.

The medium-expert and medium datasets provide broader state-action coverage than the expert datasets, contributing to improved robustness against perturbations.
This observation suggests that increasing the diversity of training datasets could enhance the robustness of offline RL methods.
To assess dataset coverage more concretely, Appendix D presents both density visualizations and a clustering-based analysis of the distribution of state-action pairs.
However, simply increasing state-action coverage through dataset augmentation may lead to low data efficiency and may even degrade policy performance.
This indicates that improving robustness may require not only offline data augmentation but also alternative strategies.

\subsection{Impact of action-perturbed datasets on offline RL robustness}
To investigate the potential for improving robustness against action-space perturbations, we train policies using the randomly and adversarially action-perturbed datasets and evaluate their robustness.
In this experiment, unlike the previous two experiments, the perturbation strength is set to $\epsilon=0.3$ for all legged robot environments, including Ant-v2.
The results are presented in Table~\ref{tab:tab3_results_perturbed_ds}. 

Compared to the expert dataset results, some results for the randomly-perturbed dataset show a slight increase in average episodic rewards under both the random and adversarial conditions. 
However, these increases are modest and cannot be conclusively interpreted as improvements in robustness.
One possible explanation is that the randomly-perturbed dataset is generated using a uniform distribution, which includes a wide range of perturbations—from very weak to strong ones. 
Such weak perturbations have little effect on the agent's behavior, but their presence in the dataset can enhance the diversity of state-action pairs, resembling those found in medium-expert or medium datasets. 
The diversity introduced by the dataset may lead to modest improvements in robustness under certain conditions. 
However, this suggests that such diversity alone may not be sufficient to improve robustness.

The results for the adversarially-perturbed dataset reveal that the average episodic rewards remain low, even under the normal condition.
These findings cast doubt on the effectiveness of adding perturbations to the training dataset for improving the testing-time robustness.

Perturbation-adding techniques have demonstrated efficacy in improving the testing-time robustness of online RL~\cite{domain_random_1,domain_random_2,adv_trng_1,adv_trng_3}.  
Online RL methods succeed because the agent can interact with the environment directly, immediately observe the rewards and next states following perturbed actions, and update the policy iteratively based on that feedback.
In contrast, offline RL enforces conservative constraints to keep the learned policy close to the dataset’s behavior distribution.
As a result, learning under action-space perturbations is only successful when the dataset includes actions already optimized for such perturbations. 
These actions must be accompanied by the corresponding rewards and next states. 
Our perturbed datasets contain only the perturbed actions, without the associated outcome information, so the policy cannot properly evaluate or learn corrective actions—hence the limited robustness gains.

As an alternative dataset modification technique, reward-based interventions such as reward shaping~\cite{rew_shape_1,rew_shape_2,rew_shape_3} or reward labeling~\cite{rew_label_1,rew_label_2} have been proposed to refine the reward signal and improve learning. These techniques have been used in offline RL, for example to denoise reward signals~\cite{rew_shape_2} or to facilitate learning in environments with sparse rewards~\cite{rew_shape_3}.
However, enhancing robustness under perturbations—as is the focus of this study—cannot be achieved solely through reward modification. In offline settings, such robustness requires not only appropriate rewards but also access to trajectories that are robust to perturbations. Since such trajectories are typically difficult to collect, reward-based approaches offer limited practical utility in this context.

As methods for achieving robust policy learning, policy ensemble methods~\cite{ensemble_1,ensemble_2,ensemble_3} have also been explored in the context of online RL. 
These methods typically aggregate multiple independently trained policies to reduce action-selection variance or stabilize policy behavior.  
However, in offline settings, policy ensembles are primarily used to mitigate issues arising from distribution shift~\cite{ensemble_off_1,ensemble_off_2,ensemble_off_3}, and they cannot overcome the fundamental limitation of offline RL—namely, the lack of online feedback.

By comparison, in hybrid offline–online approaches~\cite{hybrid_1,hybrid_2,hybrid_3,hybrid_4}, an offline-pretrained policy is finetuned through limited online interaction—allow the agent to experience real perturbations and adapt its behavior sequentially.  
This framework makes it possible to acquire corrective actions that cannot be learned from static data alone.
It therefore represents a promising direction for future work on robust offline RL.

\section{Conclusion}
This study evaluates the testing-time robustness of offline RL methods in scenarios simulating joint torque failures in legged robots. We focus on three offline RL methods---BCQ, TD3+BC, and IQL---across normal, random perturbation, and adversarial perturbation conditions. 
The results demonstrate that the offline RL methods are significantly more vulnerable to both random and adversarial perturbations compared to online RL methods in identical conditions. 
This vulnerability is attributed to the limited diversity of state-action pairs in training datasets, which restricts the adaptability of offline RL policies.
Our experiments further reveal that state-action coverage in training datasets affects the robustness of offline RL methods. Policies trained on medium-expert and medium datasets exhibit notable differences in performance under adversarial conditions when compared to those trained on expert datasets.
However, training policies on action-perturbed datasets does not yield substantial improvements in robustness, 
This result indicates that simple data augmentation alone is insufficient for achieving robustness to action-space perturbations. 
This challenge is rooted in a fundamental limitation of offline RL—datasets typically lack actions adapted to perturbed environments and their associated outcomes. As a result, the policy cannot learn to anticipate distortions that occur after action execution.
In safety-critical robotics applications—such as legged locomotion on uneven terrain or manipulation tasks near humans—actuator lag, failures and other hardware-induced problems can have severe consequences. Offline RL methods, when deployed without any form of online adaptation, cannot anticipate these post-action discrepancies. As a result, a trained policy may still fail under even slight real-world noise.
Although improved data augmentation or adaptive training strategies may contribute to robustness, their effectiveness remains limited without access to online feedback. Such feedback appears essential for reliably acquiring robustness to post-action perturbations.

To address this limitation, hybrid approaches incorporating online fine-tuning represent a promising direction. 
In particular, transitioning from offline to online reinforcement learning—by initializing policies with offline-trained models and finetuning them through limited online interactions—offers a sample-efficient means of adapting to real-world perturbations.
Online finetuning enables the agent to observe distortions occurring after action execution and to adjust the policy accordingly. Through this iterative feedback process, the agent can acquire corrective behaviors that are otherwise difficult to learn from static offline data.
    
Future work should explore such offline-to-online frameworks more deeply, with a focus on enabling robust adaptation to action-space perturbations in real-world settings. In particular, developing methods that combine the stability of offline RL with the adaptability of online finetuning will be key to deploying RL systems in the safety-critical robotics domain.

\section*{Acknowledgments}
This work was supported by JSPS KAKENHI Grant Numbers JP23K24914 and the Telecommunications Advancement Foundation.

\bibliographystyle{unsrt}  
\bibliography{refs_arXiv} 

\clearpage
\appendix

\section{Perturbation Strength Configuration}
\label{app.A}
\begin{figure*}[!t]
    \centering
    \includegraphics[width=\textwidth]{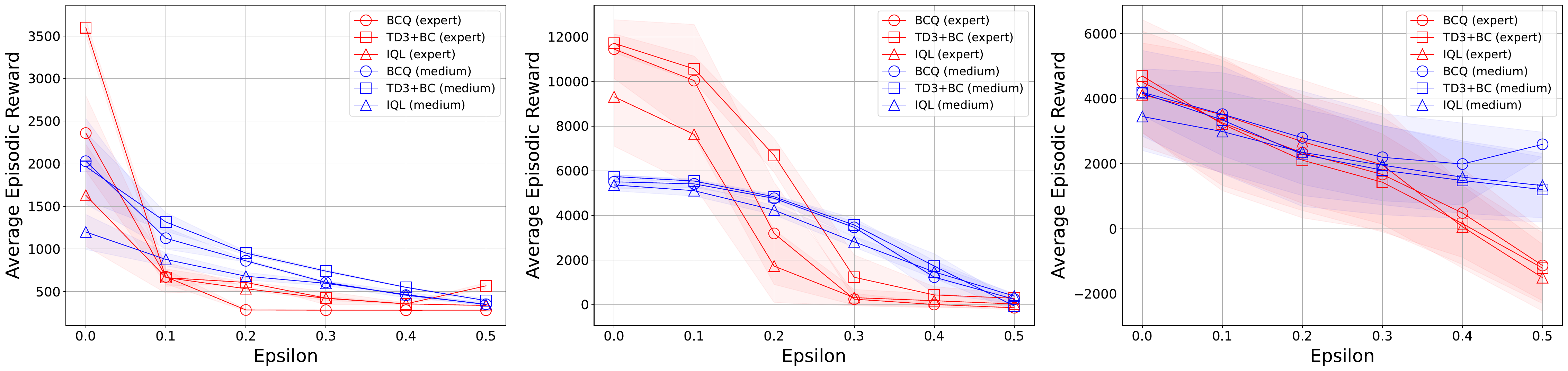}
    \caption{Testing-time robustness evaluation results under varying adversarial perturbation strengths in three legged robot environments (Hopper-v2, HalfCheetah-v2, Ant-v2; left to right). Each plot shows the average episodic rewards with shaded areas indicating standard deviation, across different perturbation levels ($\epsilon = 0.1$ to $0.5$) for three offline RL methods.}
    \label{fig:fig3_robust_eval_vs_perturbation}
\end{figure*}

To simulate worst-case scenarios, including those that might arise in real-world settings, sufficiently strong perturbation strengths are selected for each robot model.
These values were determined by empirically evaluating the performance of offline RL methods across a range of perturbation strengths, as shown in Figure~\ref{fig:fig3_robust_eval_vs_perturbation}.

Each subplot shows the testing-time robustness evaluation results under adversarial action perturbations for one of the three environments with legged robots: Hopper-v2, HalfCheetah-v2, and Ant-v2 (from left to right). 
The horizontal axis indicates perturbation strength $\epsilon$ (ranging from $0.1$ to $0.5$ in increments of $0.1$), while the vertical axis shows the average episodic reward, as defined in the \textit{Testing-time robustness evaluation} subsection of the \textit{Method} section. Each curve corresponds to a specific combination of offline RL method and dataset: BCQ (circle), TD3+BC (square), and IQL (triangle). The dataset types are color-coded: red for the expert dataset and blue for the medium dataset.

The perturbation strengths used in the main experiments were determined based on the Expert dataset. In Hopper-v2 and HalfCheetah-v2, we observed that performance degradation plateaus around $\epsilon = 0.3$, indicating that this level is sufficiently strong to induce notable adversarial impact without entirely suppressing diversity effects. Beyond this point (e.g., at $\epsilon = 0.5$), performance changes become marginal, likely due to saturation effects.
In the medium dataset, we observe a similar degradation trend between $\epsilon=0.1$ and $\epsilon=0.3$, although the overall reward values are lower and the curves exhibit slightly more gradual declines. Despite this, the relative ordering of the methods and the nature of degradation remain consistent with those in the Expert dataset. This consistency justifies the reuse of $\epsilon=0.3$ as the evaluation point.

For Ant-v2, performance under the expert dataset drops below zero at $\epsilon = 0.5$, indicating a sufficiently strong perturbation. This trend is also seen in the Medium dataset, although with a milder decline. Given the clearer degradation in Ant-v2, and guided by insights from Hopper-v2 and HalfCheetah-v2, where overly strong perturbations may obscure diversity effects, we retain $\epsilon=0.5$ as the evaluation setting for Ant-v2. 
Although perturbation strengths were selected using only the expert dataset, the consistent degradation trends observed in the medium dataset support the validity of these choices across datasets.

\begin{table}[!t]
    \caption{Hyperparameter configurations for offline and online RL algorithms. Environment abbreviations: Hopper-v2 (Hop), HalfCheetah-v2 (HC), Ant-v2 (Ant).}
    \centering
    {\begin{tabular}{cll}
        \toprule            
        Algorithm&Hyperparameter&value\\
        \midrule
        \multirow{9}{*}{BCQ}
            &Learning rate&$1\mathrm{e}{-3}$ \\
            &VAE learning rate&$1\mathrm{e}{-3}$ \\
            &Mini-batch size&$100$ \\
            &$\lambda$&$0.75$ \\
            &Hidden units&$[400, 300]$ \\
            &VAE hidden units&$[750, 750]$ \\ 
            &VAE latent size&$2 \times |\mathcal{A}|$ \\
            &Perturbation range&$0.05$ \\
            &Action samples&$100$ \\
        \midrule
        \multirow{7}{*}{TD3+BC}
            &Learning rate&$3\mathrm{e}{-4}$ \\
            &Mini-batch size&$256$ \\
            &Policy noise&$0.2$ \\
            &Policy noise clipping&$(-0.5,0.5)$ \\  
            &Policy update frequency&$2$ \\ 
            &$\alpha$&$2.5$ \\
        \midrule
        \multirow{6}{*}{IQL}
            &Learning rate&$3\mathrm{e}{-4}$ \\
            &V-function learning rate&$3\mathrm{e}{-4}$ \\
            &Mini-batch size&$256$ \\
            &Expectile&$0.7$ \\
            &Inverse temperature&$3.0$ \\   
            &Actor learning rate scheduler&Cosine \\
        \midrule
        \multirow{8}{*}{PPO}
            &\multirow{3}{*}{Learning rate}&$1\mathrm{e}{-4}$ (Hop) \\
            &&$1\mathrm{e}{-4}$ (HC) \\
            &&$3\mathrm{e}{-4}$ (Ant) \\
            &V-function learning rate&$3\mathrm{e}{-4}$ \\
            &Mini-batch size&$256$ \\
            &Clipping range&$0.2$ \\ 
            &Hidden units&$[64, 64]$ \\
            &GAE parameter&$0.95$ \\
        \midrule
        \multirow{2}{*}{SAC}
            &Learning rate&$3\mathrm{e}{-4}$ \\
            &Mini-batch size&$256$ \\
        \midrule
        \multirow{5}{*}{TD3}
            &Learning rate&$3\mathrm{e}{-4}$ \\
            &Mini-batch size&$256$ \\
            &Policy noise&$0.2$ \\
            &Policy noise clipping&$(-0.5,0.5)$ \\  
            &Policy update frequency&$2$ \\ 
        \bottomrule
    \end{tabular}}
    \label{tab:tab4_hyperparams}
\end{table}

\section{Hyperparameter configurations for offline RL and online RL algorithms}
\label{app.B}
The hyperparameters used for the offline and online RL algorithms in this study are listed in the table~\ref{tab:tab4_hyperparams} below.
Offline RL algorithms (BCQ, TD3+BC, and IQL) and online RL algorithms (SAC and TD3) are trained using the default hyperparameter settings of the d3rlpy~\cite{d3rlpy}.
The PPO hyperparameters are set according to the original paper~\cite{PPO} and the related study~\cite{ppo_param}. 
Across all algorithms, the discount factor is set to $0.99$, the target update rate is $5\mathrm{e}{-3}$, and the optimizer is Adam~\cite{Adam}. Unless otherwise specified, the default architecture is a multilayer perceptron with two hidden layers of size $[256, 256]$.

\section{Action Distributions in Action-perturbed Datasets}
\begin{figure}[t!]
    \centering
    \includegraphics[width=0.9\textwidth]{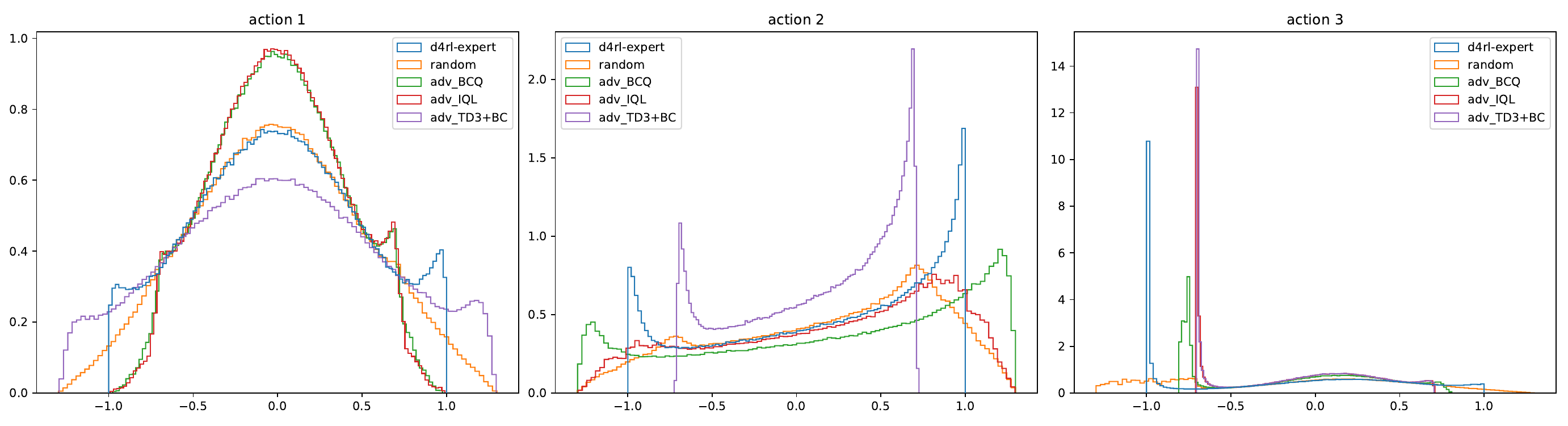} \\
    (a) Hopper-v2 \par
    \includegraphics[width=0.9\textwidth]{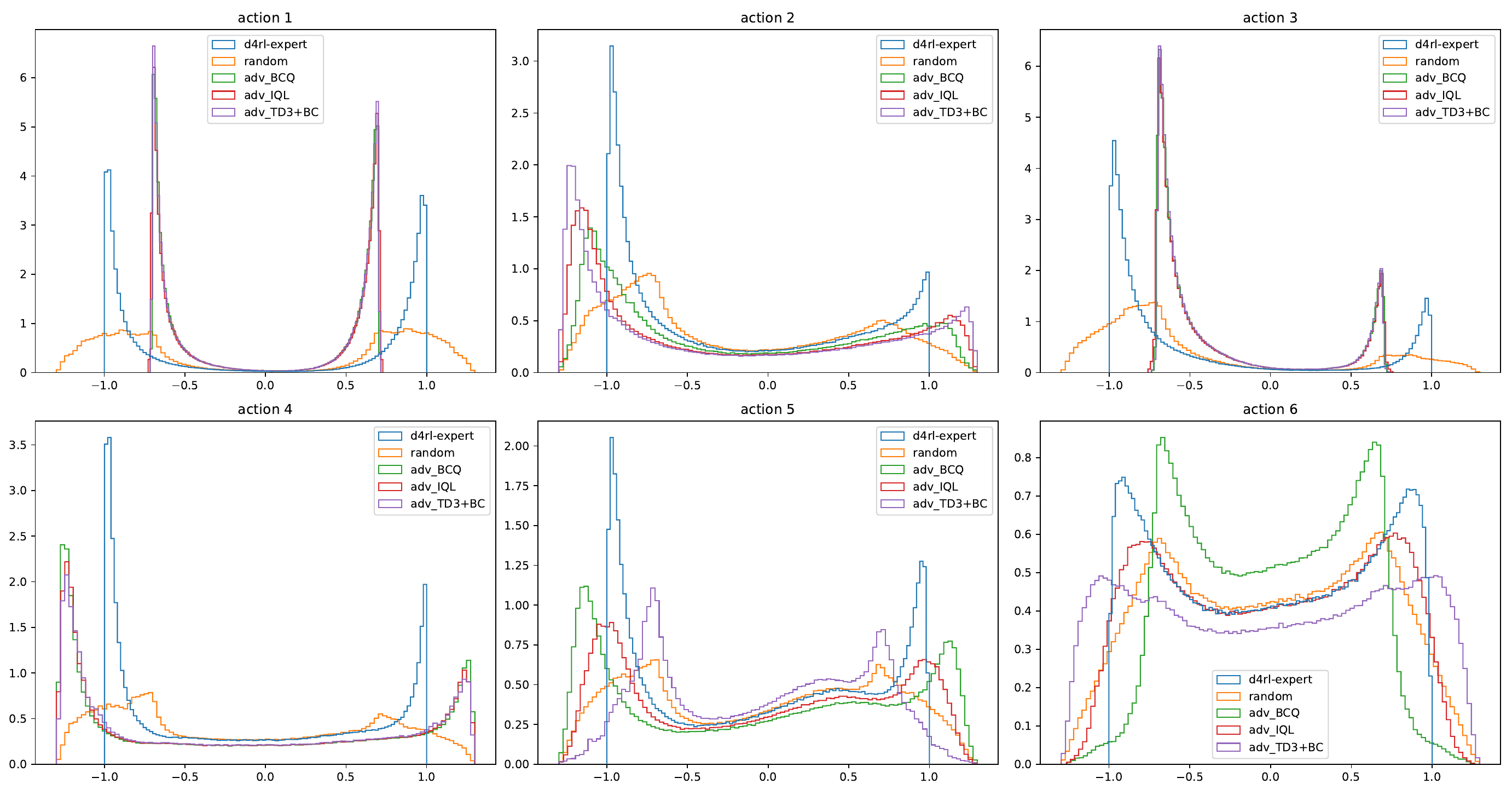} \\
    (b) HalfCheetah-v2 \par
    \includegraphics[width=0.9\textwidth]{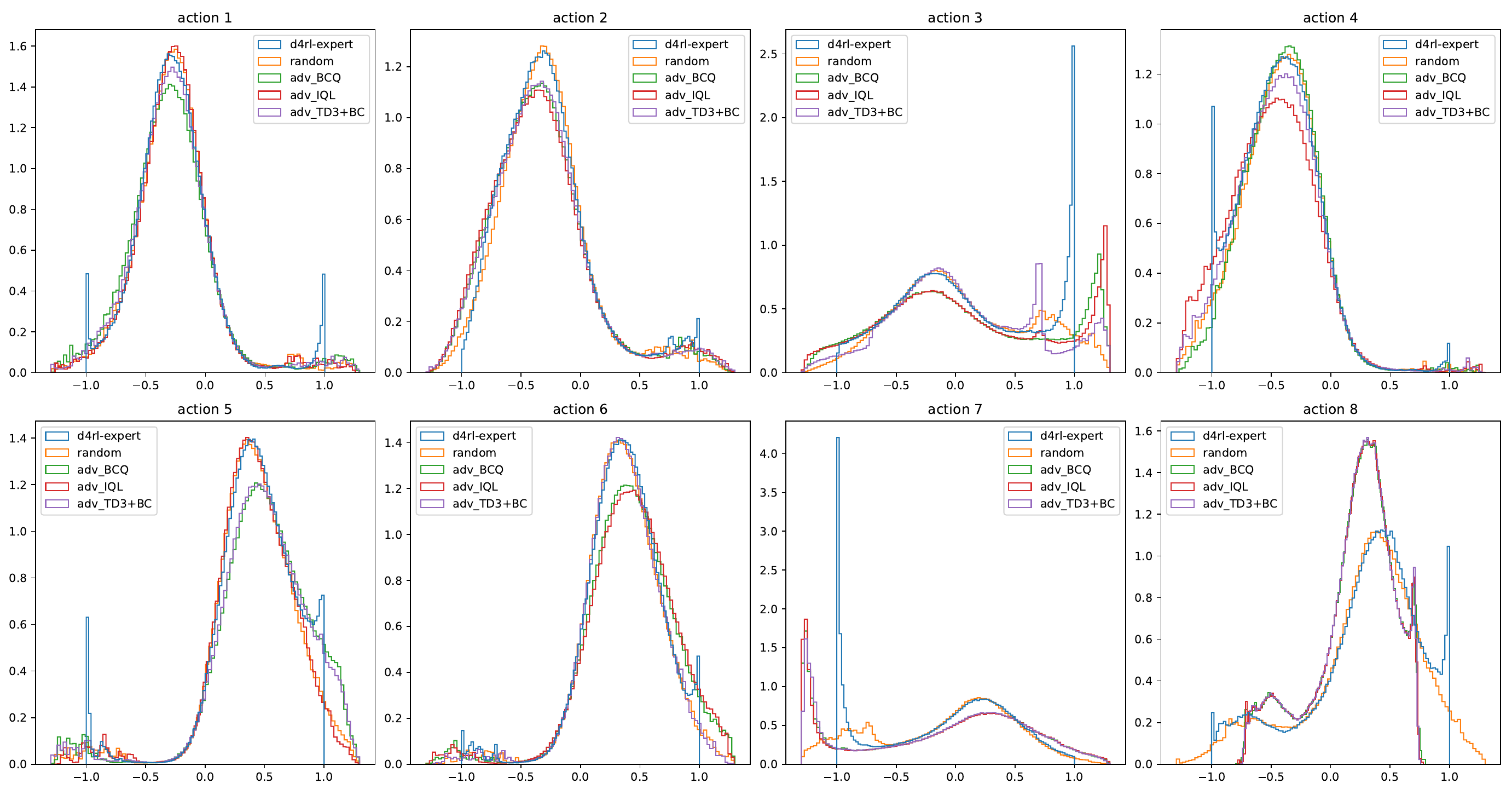} \\
    (c) Ant-v2 \par
    \caption{
        Action distributions in the training datasets for Hopper, HalfCheetah, and Ant.
        Each histogram block shows the distribution of each action dimension, comparing expert, randomly-perturbed (random), and adversarially perturbed actions (adv\_BCQ, adv\_IQL, adv\_TD3+BC).
    }
    \label{fig:fig4_perturbed_action_dist}
\end{figure}

To visualize how perturbations affect the training datasets, Figure~\ref{fig:fig4_perturbed_action_dist} shows histograms of the action values along each dimension.  
The plots include expert actions, randomly perturbed actions, and adversarially perturbed actions.  
Since adversarial perturbations are generated separately for each offline RL algorithm (BCQ, IQL, and TD3+BC), a corresponding dataset is included for each.

These histograms reflect the action distributions actually used for training and help illustrate how different perturbation strategies modify the data.  
For example, random perturbations lead to broader distributions, while adversarial perturbations introduce more targeted shifts in specific action dimensions.

\section{Dataset Coverage Analysis}
\begin{figure}[H]
    \centering
    \includegraphics[width=\textwidth]{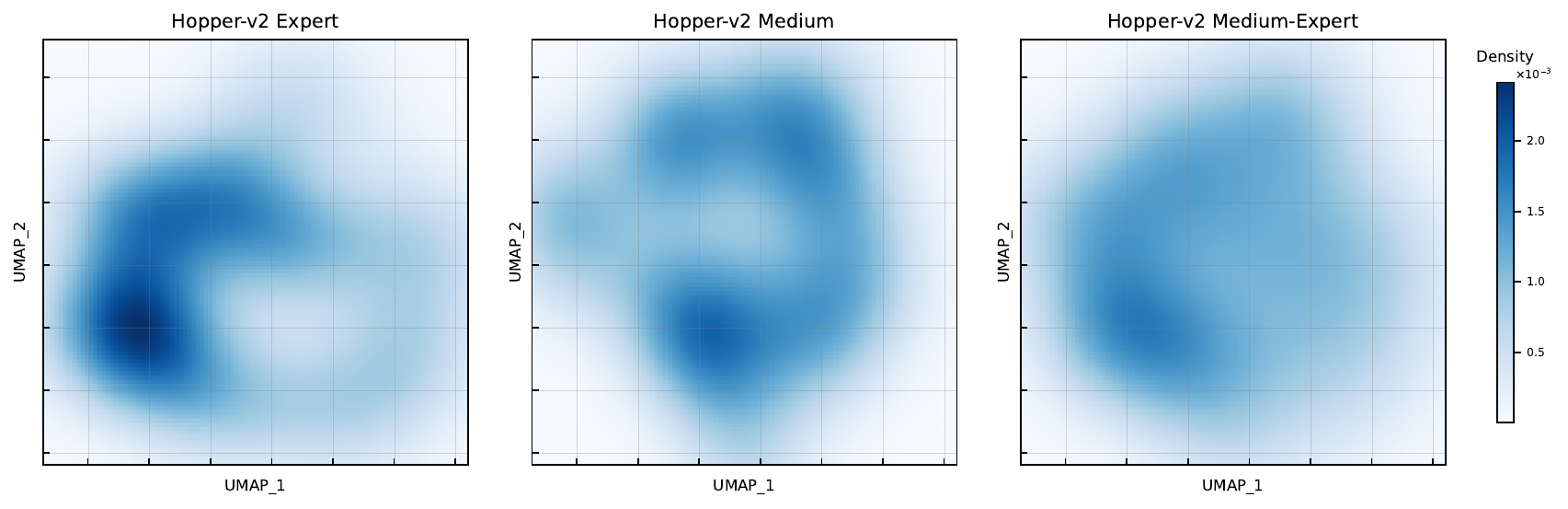}
    \includegraphics[width=\textwidth]{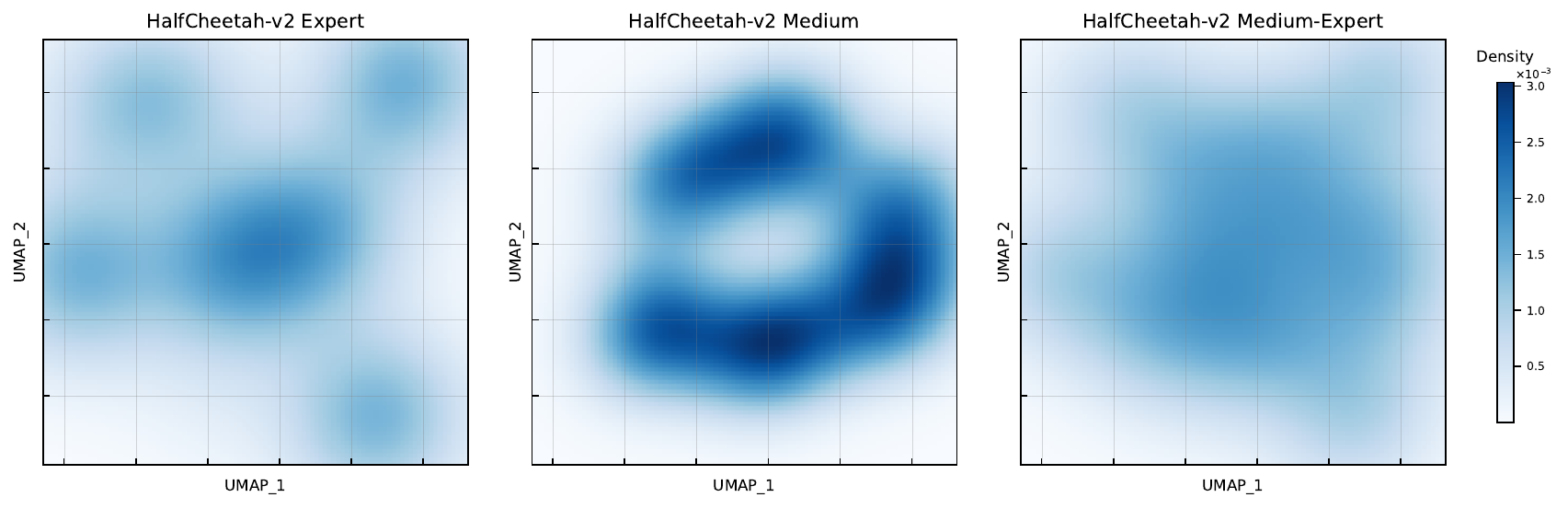}
    \includegraphics[width=\textwidth]{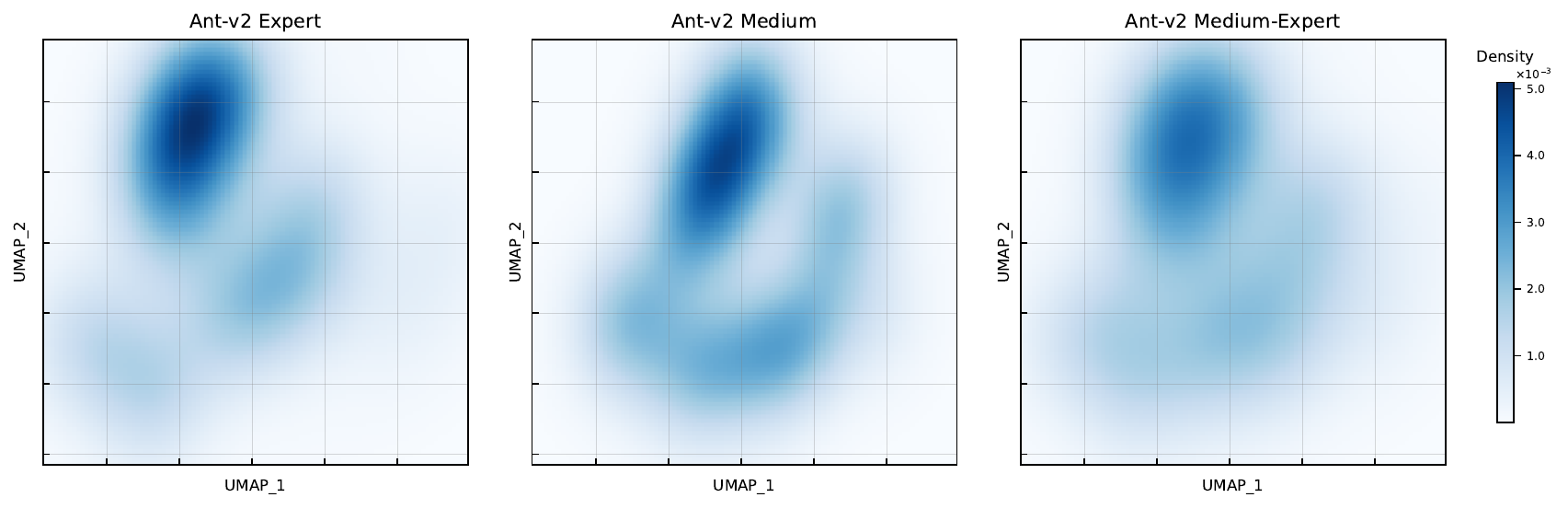}
    \caption{
    State-action coverage in the D4RL expert, medium, and medium-expert datasets across three legged robot environments.  
    Each density map is constructed by embedding scaled state-action vectors using UMAP, followed by kernel density estimation.  
    The expert datasets exhibit concentrated clusters, whereas the medium and medium-expert datasets display more dispersed coverage, indicating greater diversity.
    }
    \label{fig:fig5_state_action_density_map}
\end{figure}

To assess dataset coverage, the distribution of state-action pairs in the expert, medium, and medium-expert datasets from Hopper-v2, HalfCheetah-v2, and Ant-v2 environments is visualized, as shown in Figure~\ref{fig:fig5_state_action_density_map}. 
Each sample is first converted into a feature vector by concatenating its state and action vectors. 
To balance the influence of state and action in the embedding, each action dimension is scaled by a factor $\rho=\sqrt{d_\text{state}/d_\text{action}}$, where $(d_\text{state}, d_\text{action})=(11, 3)$ for Hopper-v2, $(17, 6)$ for HalfCheetah-v2, and $(111, 8)$ for Ant-v2. 
The corresponding values of $\rho$ are $1.915$, $1.683$, and $3.725$, respectively. 
Dimensionality reduction is performed using Uniform Manifold Approximation and Projection (UMAP)~\cite{umap}, with the number of neighbors set to $15$, minimum distance of $0.1$, and Euclidean distance as the metric. To reveal the distribution of state-action pairs in the embedded space, kernel density estimation~\cite{kde} with a gaussian kernel is applied over a $100\times100$ grid, using a bandwidth of $0.5$.
As shown in Figure~\ref{fig:fig5_state_action_density_map}, the expert datasets exhibit concentrated clusters, indicating many state-action pairs are similar and densely packed. 
In contrast, medium and medium-expert datasets show broader distributions with localized high-density regions, reflecting greater diversity in state-action coverage.

\begin{figure}[H]
    \centering
    \includegraphics[width=\textwidth]{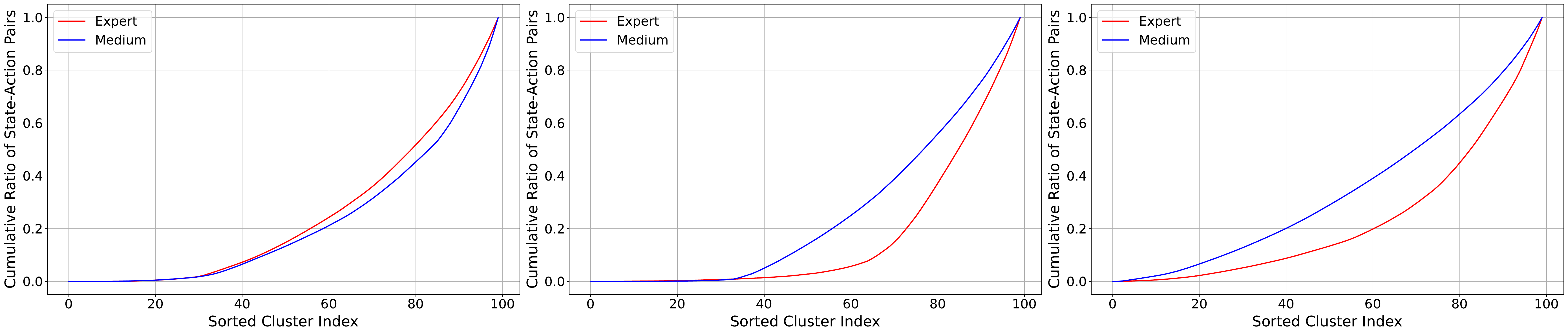}
    \caption{Cumulative ratio of state-action pairs across K-means clusters for Hopper-v2 (left), HalfCheetah-v2 (center), and Ant-v2 (right). 
    Clusters (K=100) are sorted by ascending size, and the y-axis shows the cumulative ratio of assigned samples. 
    Red and blue curves indicate expert and medium datasets, respectively. 
    Compared to the expert dataset, the medium dataset show broader distribution in Ant-v2 and HalfCheetah-v2 environments, while the two curves are nearly identical in the Hopper-v2 environment.}
    \label{fig:fig6_state_action_cumulative_ratio}
\end{figure}

In addition to the density visualizations, dataset diversity is quantitatively evaluated by analyzing how state-action pairs are distributed across clusters derived via K-means clustering. 
For each environment, a joint clustering was performed on expert and medium datasets using $K=100$ clusters, with each input represented by a concatenated state-action vector.
Figure~\ref{fig:fig6_state_action_cumulative_ratio} presents the cumulative ratio of state-action pairs with respect to cluster index. 
The x-axis shows clusters sorted in ascending order of size—that is, from clusters with the fewest to the most assigned samples—while the y-axis shows the cumulative ratio of samples.
A slower cumulative increase suggests a more even distribution over many clusters, whereas a steeper curve indicates that a large portion of the dataset is concentrated in fewer clusters.
In Ant-v2 and HalfCheetah-v2 environments, the medium datasets display smoother cumulative growth, indicating broader coverage of the state-action space.
In contrast, the expert datasets show steeper increases beyond a certain point, indicating repeated visitation of similar state-action pairs.
In the Hopper-v2 environment, both curves nearly coincide, reflecting similar diversity in expert and medium trajectories.
This analysis reinforces the hypothesis that broader coverage contributes to robustness, and is consistent with the experimental finding that performance gains from medium datasets are more improved in Ant-v2 and HalfCheetah-v2 than in Hopper-v2.

\clearpage

\clearpage

\end{document}